\documentclass[final]{cvpr}

\usepackage{times}
\usepackage{epsfig}
\usepackage{graphicx}
\usepackage{amsmath}
\usepackage{amssymb}
\usepackage{enumerate}
\usepackage{booktabs} 
\usepackage{array}  %
\usepackage{multirow}
\usepackage[tight,scriptsize,sf,SF]{subfigure}
\usepackage{algorithm}
\usepackage{algorithmic}
\usepackage{pifont}
\newcommand{\cmark}{\ding{51}}%
\newcommand{\xmark}{\ding{55}}%

\usepackage[pagebackref=true,breaklinks=true,colorlinks,bookmarks=false]{hyperref}

\def\cvprPaperID{****} 
\def\confYear{CVPR 2021}

\begin{document}

\title{Real-Time and Accurate Object Detection in Compressed Video by Long Short-term Feature Aggregation}

\author{Xinggang Wang$^{1}$, Zhaojin Huang$^{1}$, Bencheng Liao$^{1}$, Lichao Huang$^{2}$, Yongchao Gong$^{2}$, Chang Huang$^{2}$, \\
$^{1}$Huazhong University of Science and Technology,\\
$^{2}$Horizon Robotics Inc.  \\
    {\tt\small \{xgwang, zhaojinhuang,lbc\}@hust.edu.cn \{lichao.huang,yongchao.gong,chang.huang\}@horizon.ai}
    
}
\maketitle
\begin{abstract}
Video object detection is a fundamental problem in computer vision and has a wide spectrum of applications. Based on deep networks, video object detection is actively studied for pushing the limits of detection speed and accuracy. To reduce the computation cost, we sparsely sample key frames in video and treat the rest frames are non-key frames; a large and deep network is used to extract features for key frames and a tiny network is used for non-key frames. To enhance the features of non-key frames, we propose a novel short-term feature aggregation method to propagate the rich information in key frame features to non-key frame features in a fast way. The fast feature aggregation is enabled by the freely available motion cues in compressed videos. Further, key frame features are also aggregated based on optical flow. The propagated deep features are then integrated with the directly extracted features for object detection. The feature extraction and feature integration parameters are optimized in an end-to-end manner. The proposed video object detection network is evaluated on the large-scale ImageNet VID benchmark and achieves 77.2\% mAP, which is on-par with the state-of-the-art accuracy, at the speed of 30 FPS using a Titan X GPU. The source codes are available at \url{https://github.com/hustvl/LSFA}.

\end{abstract}

\section{Introduction}

Object detection in still images has made great progress based on deep learning. However, in the case of video object detection, directly applying still image object detectors to each individual video frame causes redundant computational cost and requires post-processing to retrieve temporal context for higher detection accuracy. Consequently, a principled framework that can seamlessly model temporal information among the video and provide real-time detection speed is of urgent demand.

Current video object detection works focus on improving the accuracy and speed from the following perspectives. The techniques for higher detection accuracy are performed either on box-level ~\cite{kang2018t, han2016seq, chen2018optimizing, feichtenhofer2017detect} or feature-level ~\cite{zhu2017flow, zhu2017deep, zhu2018towards, wang2018fully, hetang2017impression, wang2019learning}. The box-level methods aim at associating the boxes of the same object and re-scoring the detection hypotheses using the contextual information in temporal dimension, while the feature-level methods target at obtaining more powerful deep feature maps for each frame by fusing the features from nearby frames. For higher detection speed, the main observation is that there is lots of redundant information in successive frames so that it is not necessary to spend equally expensive computation for them. These methods usually divide the video frames into key frames and non-key frames. They use a large network to extract the features of the key frames and propagate these features to non-key frames, e.g. ~\cite{zhu2017deep}. The major issues required to be addressed by these speeding-up methods lie in the large computation cost of feature propagation and the compensation for the quality loss of propagated features for non-key frames.

In this paper, we develop a real-time video object detection framework based on feature propagation with improved object detection accuracy. The core idea of this framework, named \textit{long short-term feature aggregation} (LSFA). For the given key frames and non-key frames in a video, we perform long-term and short-term feature aggregation on them, respectively. Specifically, the key frame feature extracted by a large deep network and it is aggregated with previous key frame features guided by optical flow between two key frames. This aggregation provides long-term information that alleviates the problems of motion blur, partial occlusion, viewpoint variation etc., yielding more robust features for the key frames. Once a key frame feature is obtained, it can be used for estimating the features of its subsequent non-key frames via short-term feature propagation. Specifically, this short-term feature propagation is guided by the motion vector and residual error information rather than optical flow, because the motion vector and residual error information are freely available in compressed video~\cite{wu2018compressed}, i.e., no computation cost is needed. In the experiments, we find that the motion vector and residual error are on par with optical flow for feature propagation from key frames to non-key frames. However, the non-key frame features obtained by short-term feature propagation lack the information directly extracted from images and they inevitably suffer from low quality. To address this issue, we propose short-term feature aggregation by integrating the propagated feature with the feature extracted by a tiny network. Thus, we obtain reliable non-key frame features in a fast way.

When implementing the proposed LSFA framework, we keep optimizing the speed of detection. Short-term feature aggregation is performed more frequently than long-term feature aggregation, since the number of non-key frames is much larger than the number of key frames. Thus, we use motion vector which can cheaply extracted from compressed video ~\cite{wu2018compressed, wang2018fodcv}, instead of the widely used FlowNet~\cite{dosovitskiy2015flownet}, to guide feature propagation between key frame and non-key frame. For feature extraction from non-key frames, we utilize a tiny network which only consists of the first two blocks in ResNet~\cite{he2016deepresnet}, with downsampled input (the image is downsampled for four times). In addition to the propagated and extracted features, we find the feature learned from the residual errors is also helpful. Therefore, the short-term feature aggregation also contains the residual error feature. For long-term feature aggregation, we follow the sparsely recursive feature aggregation (SRFA) scheme in ~\cite{zhu2018towards}. The feature propagation between sparse key frames is guided by the optical flow provided by FlowNet, as there is no motion vector between them. It should be noted that long-term feature aggregation not only provides robust features for key frames, but also improves the robustness of non-key frames, as the non-key frame features are partially inherited from them. 
Compared with ~\cite{zhu2018towards} that only performs feature propagation based on FlowNet for non-key frames, our method largely benefits from the novelly proposed short-term feature aggregation. The main advantages of our method mainly lie in two aspects: (1) It utilizes the information from small non-key frames which helps to improve the quality of the propagated feature. (2) It takes advantage of motion vector that is obtained for nearly free without the need of computing optical flow and the motion vector has been verified as accurate as the flow produced by FlowNet in our experiments. 

Our main contributions can be summarized as follows:
\begin{enumerate}
\item We propose a unified framework named LSFA for video object detection addressing both detection accuracy and speed.
\item In LSFA, the short-term feature aggregation method is the first work that uses feature extracted from original image to enhance the propagated features for non-key frames.
\item On the challenging ImageNet VID dataset, LSFA runs in real-time (30 FPS) with detection accuracy on-par with the state-of-the-art method (77.2\% mAP vs 77.8\% mAP in ~\cite{zhu2018towards}). With the strong performance, we believe LSFA is helpful for various practical applications. 
\end{enumerate}

In the rest of this paper, Section~\ref{sec:rw} reviews related work, Section~\ref{sec:met} presents the our main method, i.e., long short-term feature aggregation for video object detection, Section~\ref{sec:exp} conducts experiments on the ImageNet VID dataset, and lastly Section~\ref{sec:con} concludes the paper.

\section{Related Work}
\label{sec:rw}
\paragraph{Object detection in still images}

 Deep learning has made great progress in object detection in still images ~\cite{ren2015faster,dai2016rfcn,he2017maskrcnn,huang2019msrcnn,redmon2016youyolo,liu2016ssd, hou2018object,  nie2019enriched, wang2019learning, tian2019fcos}. There are single-stage methods, such as YOLO~\cite{liu2016ssd} and two-stage methods, such as Faster RCNN~\cite{ren2015faster}, R-FCN~\cite{dai2016rfcn} and Mask RCNN~\cite{he2017maskrcnn}. In this paper, we build our video object detector based on the two-stage methods for two reasons: (1) currently nearly all the Video Object Detection (VOD) research works are based on the two-stage methods; and (2) the two-stage detectors have better detection accuracy and lower speed compared to the single-stage detectors, that means improving the speed two-stage detectors for VOD is more urgent. Specifically, our method is based on R-FCN \cite{dai2016rfcn}, which is more efficient and effective than Faster R-CNN. However, detecting each frame in video individually will bring redundant computational cost. In the proposed video object detection method, non-key frame features are extracted based on key frame features to avoid the computation redundancy.

\paragraph{Video object detection}

Current Video Object Detection (VOD) research is benchmarked by the VID dataset introduced by ILSVRC~\cite{ILSVRC15} in year 2017. Exiting video object detection algorithms can be divided into two streams. One is box-level method, the other is feature-level method. We will introduce them in detail as follows.

Box-level VOD methods mainly focus on linking box from different frames. T-CNN~\cite{kang2018t} warps the detected boxes in previous frame to the current frame, and generates tubelets by tracking the boxes across different frames. The boxes along each tubelet will be re-scored based on the tubelet classification result. Seq-NMS~\cite{han2016seq} builds high-confidence box sequences according to boxes overlapping  and rescores the boxes in a sequence to the average or maximum confidence. D\&T ~\cite{feichtenhofer2017detect} learns an ROI tracker along with detector. The cross-frame tracker is used to boost the scores for positive boxes. ST-Lattice ~\cite{chen2018optimizing} sparsely performs expensive detection and propagates the detected boxes across both scales and time with substantially cheaper networks by exploiting the strong correlations among them. 
Most box-level VOD methods improves the detection accuracy in a post-hoc way. Now most of the methods do not work in an online fashion, i.e., computing the boxes for $t$-th frame relies the information in $(t+x)$-th ($x>0$) frames.


Feature-level VOD methods usually use the motion information between two frames to obtain feature-map-level spatial correspondence. FGFA~\cite{zhu2017flow} warps the features of nearby frames according to the correspondence provided by FlowNet~\cite{dosovitskiy2015flownet}, and aggregates the wrapped features to enhance the quality of current frame features for higher detection accuracy. DFF~\cite{zhu2017deep} equally samples key frames in video and extract key frame features using a deep network. For non-key frames, their features are directly warped from key frame features without using a deep feature extraction network. 
To improve DFF, ~\cite{zhu2018towards} performs recursive feature aggregation for key frames and partial feature updating for non-key frames. 
MANet~\cite{wang2018fully} jointly calibrates the features of objects on both pixel-level and instance-level in a unified framework for better feature correspondence. 
~\cite{guo2019pro} has a similar idea with our method. We both focus on warping the feature from key frame to non-key frame. However, ~\cite{guo2019pro} establishes the spatial correspondence between features across frames in a local region with progressively sparser stride, while we obtain the motions from compressed video without any calculation directly. Besides, the tiny network in ~\cite{guo2019pro} is used for calculating feature correspondence while we use it for enhancing non-key frame features.

\paragraph{Temporal information for video object detection}


The temporal information in video provides the feasibility for improving video object detection speed. To utilize the temporal information, the feature-level VOD networks propagate information in nearby frames. According to the way of feature propagation, they can be divided into two streams: correspondence-based methods and correspondence-free methods. 

The correspondence-based methods explicitly compute the spatial correspondence across frames and wrap features. As described above, many of them use FlowNet to computer optical flow to obtain feature correspondence, e.g., DFF and FGFA. 
Besides optical flow, ~\cite{feichtenhofer2017detect} uses RoI correlation to obtain object-level correspondence. ~\cite{bertasius2018object} learns deformable offsets between two frames and uses deformable convolution for warping features.
Rather than re-computing, the correspondence can be obtained in encoded video. For example, ~\cite{chen2018optimizing} adopts motion history image (MHI) for spatial correspondence for warping features, and ~\cite{wang2018fodcv} proposes to use motion vector from compressed video for warping features. Among optical flow, MHI, motion vectors, the last one is the most computational cheapest one. Thus, we adopt motion vectors in our VOD network.

The correspondence-free methods do not need spatial correspondence and usually aggregate features using a network. For example, ~\cite{liu2018mobile} uses LSTM to integrate the previous frame features with current frame features and ~\cite{li2018low} proposes to propagate the features by spatially variant convolution. 

\begin{figure*}[htp]
\centering
\includegraphics[width=0.8\linewidth]{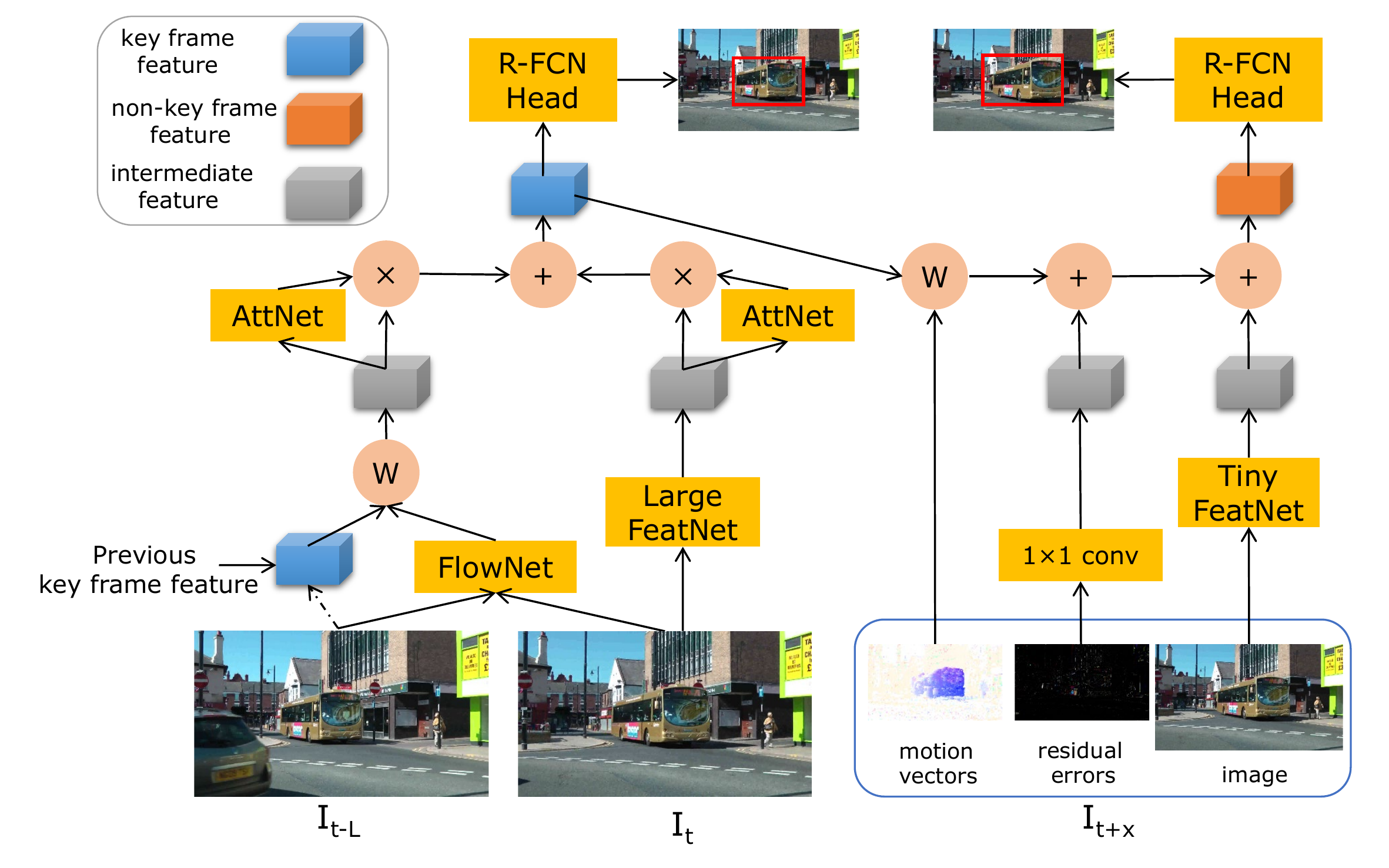}
\caption{
The overall network architecture of LSFA. The left part shows the long-term feature aggregation, and the right part shows the short-term feature aggregation. In the left part, the deep feature of a key frame (i.e., $I_t$) is obtained by attentively integrating two kinds of features: The feature extracted from a large feature extraction network, and the warped feature of its previous key frame (i.e., $I_{t-L}$) guided by FlowNet. In the right part, the feature of a non-key frame (i.e., $I_{t+x}$) is obtained via aggregating three kinds of features: The propagated feature from its previous key frame guided by motion vectors, the feature mapped from residual errors using a $1\times 1$ convolution, and the feature extracted from the downsampled image using a tiny network. On the top of the feature maps obtained by long short-term feature aggregation, the head detector of R-FCN is applied to accomplish the video object detection task. Best viewed in color.
}
\label{fig:network}
\end{figure*}

\section{Long Short-term Feature Aggregation}

\label{sec:met}
In this section, we present the details of the proposed long short-term feature aggregation (LSFA) method for video object detection. Subsection~\ref{subsec_3_1} provides the background knowledge of information extraction from compressed video, which is the basic of LSFA. Then, we introduce long-term and short-term feature aggregation in Subsection~\ref{subsec_3_2} and Subsection~\ref{subsec_3_3}, respectively. 

\subsection{Information extraction from compressed video}
\label{subsec_3_1}

Video compression algorithms, such as H.264 and MPEG-4, are used to store video frames in a highly efficient way. For each clip in a video, since the successive frames are highly similar, we can just store the first frame (called key frame) and the differences between this frame and its next frames (called non-key frames), instead of storing each frame. Most video compression algorithms divide a video into I-frames (intra-coded frames), P-frames (predictive frames), B-frames (bi-directional frames). In this paper, we only use I-frames and P-frames. I-frame is a regular image without compression, and P-frame stores the difference between the current frame and its previous frame. The differences between frames are called motion vectors. For the current frame, it can be predicted using its previous frame and motion vectors. Moreover, P-frame also contains the residual errors between the predicted image and the corresponding original image.
Therefore, the current frame can be reconstructed using motion vectors, residual errors and its previous frame. 
In video compression, the video frames are divided into different groups (i.e., clips), each containing \text{L} frames. The first frame in a group is I-frame, used as key frame, while the rest are P-frames, used as non-key frames. This means that the key frames are fixed and there is a key frame in every \text{L} continuous frames. 
It should be noted that motion vectors only describe the motion between two adjacent frames, but key frames and non-key frames may not be adjacent. For this reason, we need to calculate the motions between a key frame and a non-key frame. In our experiments, the motion vectors are obtained using the Coviar method ~\cite{wu2018compressed}.
To clearly describe the proposed method, we only need to define the following three frames:
\begin{itemize}
    \item $I_t$ denotes a key frame;
    \item $I_{t-L}$ is the previous key frame of $I_t$;
    \item $I_{t+x}$ is a non-key frame whose previous key frame is $I_t$.
\end{itemize}
With slightly abuse notations, $L$ means the length of a video segment and $x$ means the order of $I_{t+x}$ after $x_t$. Moreover, we define the motion vector map and residual error map along with $I_{t+x}$ as $M_{t+x}$ and $R_{t+x}$, respectively.


As mentioned above, motion vectors and residual errors from P-frame only describe the motions between adjacent frames, but a key frame and a non-key frame may not be adjacent. As a result, we cannot directly obtain $M_{t+x}$ and $R_{t+x}$ from $I_t$ and only a single P-frame. To tackle this problem, we trace all motion vectors back to the corresponding key frame and accumulate the residuals on the way according to ~\cite{wu2018compressed}. In this manner, we obtain the motions between the key frame and non-key frames even if they are not adjacent to each other. In short-term feature aggregation, motion vectors are used to replace the the flow calculated by FlowNet to speed up the progress of video object detection, and residual errors are also adopted for boosting the detection accuracy. 

Here we explain the reason that motion vectors and residual errors can be obtained nearly free. Video object detection can be roughly divided into two categories: off-line methods and on-line methods. For off-line video object detection, most methods first decode the video into frame-by-frame RGB images, which are then used for detection. An RGB image is decoded from I-frame and P-frame (note that P-frame contains motion vectors and residual errors). That means we can directly get motion vectors and residual errors from P-frames without any calculations. For on-line video object detection, there are mainly two cases. One is that the video is captured by a terminal camera and streams back to the server, and then object detection is performed in the server. The other is that the terminal camera performs object detection locally. For the first case, as the information from the terminal to the server is the compressed video information, it is the same as off-line video object detection, where the RGB images, motion vectors and residual errors can be obtained almost at the same time. For the other case, the camera captures RGB images directly. At this time, the motion vectors and residual errors require to be calculated separately, which will take extra time. Generally speaking, under the corresponding conditions (i.e., off-line and information streaming from terminal to server), we can obtain motion vectors and residual errors for nearly free.

\subsection{Long-term feature aggregation}
\label{subsec_3_2}

In LSFA, the proposed long-term feature aggregation is performed among key frames. It mainly consists of two parts: feature extraction and feature aggregation. For two successive key frames, e.g., $I_{t-L}$ and $I_t$, we extract their features using a large feature extraction network $\mathcal{N}_\text{large}$. Their feature maps are denoted by  $F_{t-L}$ and $F_t$ respectively, which are computed as follows,
\begin{equation}
    \begin{split}
    F_{t-L} & = \mathcal{N}_\text{large}(I_{t-L}) \\
    F_t  & = \mathcal{N}_\text{large}(I_t).
    \end{split}
\end{equation}
$F_{t-L}$ and $F_t$ have the same shape, denoted as $F_{t-L} \in \mathcal{R}^{c\times h\times w}$ and $F \in \mathcal{R}^{c\times h\times w}$, where $w$, $h$ and $c$ denote feature width, height and channel numbers respectively.

The large network for key frame feature extraction is important, because it affects not only the feature quality of key frames, but also the non-key frame detection as the non-key frame features rely on key frame features. In our experiments, we adopt the state-of-the-art ResNet-101 model ~\cite{he2016deepresnet} as the large feature extraction network. ResNet-101 is pre-trained on the ImageNet dataset. We modify ResNet-101 following DFF. Specifically, the last 1000-way classification layer is discarded, and the first block of the \textit{conv5} layers are modified to have a stride of 1 instead of 2. Moreover, we also apply the holing algorithm ~\cite{chen2014semantic} to all the $3\times 3$ convolutional kernels in \textit{conv5} to keep the field of view (dilation=2). A randomly initialized $3\times 3$ convolution is appended to \textit{conv5} to reduce the feature channel dimension to 1024, where the holing algorithm is also applied (dilation=6).

Long-term feature aggregation is based on optical flow, since there is no motion information encoded in compressed video between pairs of key frames. Following the popular setting in ~\cite{zhu2017deep}, we compute optical flow between key frames using FlowNet ~\cite{dosovitskiy2015flownet}. The computed optical flow from $I_{t-L}$ to $I_t$ is denoted by
\begin{equation}
        O_t = \mathcal{N}_\text{flow}( I_{t-L}, I_t ),
\end{equation}
where $\mathcal{N}_\text{flow}$ is a pre-trained FlowNet. Note that it is not necessary to add subscript $t-L$ in $O_t$, as the flow only has a single direction. Now, we define the long-term aggregated feature $F'_t$ for key frame $I_t$ as
\begin{equation}
\label{eq:lfa}
\begin{split}
    F'_{t} = F_t \odot A_{t} + \mathcal{W}\left( F'_{t-L}, O_t \right) \odot A_{t-L},
\end{split}
\end{equation}
where $\mathcal{W}$ is the bilinear warping function applied on all the locations for each channel in the feature maps, and $A_{t-L}$ and $A_{t}$ are the weight matrices of features $F_{t-L}$ and $F_{t}$ learned by an attention network.
$A_{t}$ and $A_{t-L}$ have the same shape of $1\times h \times w$ and operator $\odot$ performs position-wise production. Function $\mathcal{W}\left( F'_{t-L}, O_t \right)$ warps the previous aggregated key frame feature $F'_{t-L}$ to current key frame. It should be noted that Eq.~\eqref{eq:lfa} is a recursive definition as it contains $F'_{t-L}$ which is the long-term aggregated feature of $I_{t-L}$. 
The attention networks are useful to solve the problem caused by inaccurate flow. There is no guarantee that FlowNet can provide perfect pixel correspondences between key frames. The attention networks are helpful to get rid of inaccurate/false features bring by inaccurate optical flow. Specifically, if the correspondence from optical flow is inaccurate, the attention on the corresponding positions would be low and thus the inaccurate features will be ignored.

The weight matrices $A_{t}$ and $A_{t-L}$ are derived from $F_{t}$ and $F'_{t-L}$ as follows,
\begin{equation}
\label{eq:weightt}
A_{t}, A_{t-L} = \text{softmax}\left( \mathcal{N}_\text{att} \left(F_{t} \right), \mathcal{N}_\text{att} \left( \mathcal{W}\left( F'_{t-L}, O_t \right) \right) \right),
\end{equation}
where $\mathcal{N}_\text{att}$ is an attention network. 
Both $\mathcal{W}\left( F'_{t-L}, O_t \right))$ and $F_t$ describe the visual context of the same frame, thus we learn the corresponding weight matrices from the two features to attentively integrate them. $\mathcal{N}_\text{att}$ has three layers, namely a $3\times 3 \times 256$ convolution, a $1 \times 1 \times16$ convolution and a $1 \times 1 \times 1$ convolution, which are randomly initialized before training. 

In Fig.~\ref{fig:network}, the left part illustrates the long-term feature aggregation, which exactly corresponds to Eq.~\eqref{eq:lfa}, i.e., \texttt{AttNet}, \texttt{FlowNet} and \texttt{Large FeatNet} correspond to $\mathcal{N}_\text{att}$, $\mathcal{N}_\text{flow}$ and $\mathcal{N}_\text{large}$ respectively.  

\subsection{Short-term feature aggregation}
\label{subsec_3_3}

Short-term feature aggregation is the core contribution of this work. In this part, we are the first to propose to use the extracted feature from the original image to enhance the propagated feature for a non-key image. Its graphical illustration is in the right part in Fig.~\ref{fig:network}. As demonstrated, the desired non-key frame feature is composed by the propagated feature guided by motion vector, the intermediate feature from residual error map and the feature extracted from a downsampled frame, formulated as follows,
\begin{equation}
\label{eq:sfa}
\begin{split}
    F'_{t+x} & = \mathcal{W}\left( F'_t, M_{t+x} \right)    \\
             & + \text{Conv}\left( R_{t+x} \right)  \\
             & + \mathcal{N}_\text{tiny} \left( D\left( I_{t+x} \right)  \right).
\end{split}
\end{equation}

The first item in Eq.~\eqref{eq:sfa} denotes the feature propagated from the previous key frame feature, which is obtained via long-term feature aggregation. The propagation is guided by the motion vector $M_{t+x}$ extracted from compressed video, instead of FlowNet. 
In our implementation, the width and height of $F'_t$ are both $1/16$ of the original image width and height respectively, thus the motion vector map is correspondingly downsampled via bilinear interpolation to match the size of  $F'_t$. 
As will be verified in our experiments, using motion vectors to guide the feature warping significantly improves the speed of video object detection.  

The second term in Eq.~\eqref{eq:sfa} extracts feature from the residual error map simply using a $1\times 1$ convolution layer. Since the residual errors  are already quite informative, we empirically find it is unnecessary to use any multiple layers or deep networks for further feature extraction. Similar to the case of motion vector, $R_{t+x}$ is also downsampled via bilinear interpolation, and then a $1\times 1$ convolution is adopted to ensure that $\text{Conv}\left( R_{t+x} \right)$ has the same channel number with $F'_t$.

The third item in Eq.~\eqref{eq:sfa} indicates the feature extracted from the original image of non-key frame $I_{t+x}$. In order to make the feature extraction faster, we downsample $I_{t+x}$ by a factor of $1/4$, and denote it as $D\left( I_{t+x} \right)$. Afterwards, we apply a tiny feature extraction network $\mathcal{N}_\text{tiny}$ to it. $\mathcal{N}_\text{tiny}$ consists of the first two blocks (\textit{conv1}, \textit{conv2}) of ResNet-101~\cite{he2016deepresnet} and a $3\times3$ convolution. The parameters of the two blocks are initialized as the values in ResNet-101 pretrained on ImageNet. The $3\times3$ convolution is randomly initialized. The purpose of the $3\times3$ convolution is to both match the shape of $F'_t$ and adapt the pretrained feature to the propagated feature. Furthermore, as we downsample the non-key frame, $\mathcal{N}_\text{tiny} \left( D\left( I_{t+x} \right)  \right)$ can also provide multi-scale information, which is also useful for object detection. Based on the above designs, the computational cost of $\mathcal{N}_\text{tiny} \left( D\left( I_{t+x} \right)  \right)$ is negligible compared with that of feature extraction for a key frame, i.e. $\mathcal{N}_\text{large}(I_t)$.

\subsection{Training}

In the training phase, we need three kinds of frames: previous key frame, current key frame, current (non-key) frame. We first use previous key frame and current key frame for long-term feature aggregation. After getting the aggregated feature $F'_t$, we use motion vectors to warp $F'_{t}$ to the current frame, and then fusing with the residual errors and the feature extracted by the tiny network. The large and tiny feature networks, attention networks, R-FCN heads are end-to-end trained. The FlowNet is a pre-trained model and its parameters do not require to be trained.

\subsection{Inference}
Algorithm~\ref{alg1} summarizes the inference algorithm. We split the frame sequence into segments of equal length $L$. The first frame of each segment is used as the key frame, and the rest are non-key frames. Lines 1$\rightarrow$2 describe the initialization of the inference phase. Lines 4$\rightarrow$10 introduce the steps of key frame inference. 
We use $\mathcal{N}_\text{large}$ to extract feature $F_t$ from an input frame. The previous key frame feature is warped to the current key frame by FlowNet ($\mathcal{N}_\text{flow}$) first, with the calculated flow denoted by $O_t$. We then fuse the feature $F_t$ with the feature warped from previous key frame feature $F'_{t-L}$ using $O_t$. Moreover, an attention network $\mathcal{N}_\text{att}$ is adopted to learn the weights of these two features, and then integrate the features using the weights. Lines 12$\rightarrow$14 present the inference of non-key frames. The key frame feature is warped by the motion vector ($M_{t+x}$) and then added to the residual error which has gone through a $1\times1$ convolution, as well as the feature from the tiny network. For both key frame and non-key frames, we all use the R-FCN head (denoted by $\mathcal{N}_\text{head}$) to produce the detection results.

\begin{algorithm}
	\caption{Inference Procedure of LSFA} 
	\label{alg1}
	\begin{algorithmic}[1]
		\REQUIRE video frames $\left\{I_i \right\}$,  segment length L
		\ENSURE  detection results $\left\{y_i \right\}$
		\STATE $F'_0=\mathcal{N}_\text{large}(I_0)$
		\STATE $y_0=\mathcal{N}_\text{head}(F'_0)$
		\FOR{$i = 1 \rightarrow	 \infty$} 
		\IF {$i \mod L = 0$ }  
		\STATE $t=i$
		\STATE $F_t=\mathcal{N}_\text{large}(I_t)$
		\STATE $O_t = \mathcal{N}_\text{flow}( I_{t-L}, I_t )$
		\STATE $A_{t}, A_{t-L} = \text{softmax}\left( \mathcal{N}_\text{att} \left(F_{t} \right), \mathcal{N}_\text{att} \left( \mathcal{W}\left( F'_{t-L}, O_t \right) \right) \right)$  
		\STATE $F'_{t} = F_t \odot A_{t} + \mathcal{W}\left( F'_{t-L}, O_t \right) \odot A_{t-L}$
		\STATE $y_t=\mathcal{N}_\text{head}(F'_t)$
		\ELSE  
		\STATE $x=i-t$
		\STATE  $F'_{t+x} = \mathcal{W}\left( F'_t, M_{t+x} \right) + \text{Conv}\left( R_{t+x} \right) +  \mathcal{N}_\text{tiny} \left( D\left( I_{t+x} \right) \right)$ 
		\STATE $y_{t+x}=\mathcal{N}_\text{head}(F'_{t+x})$
		\ENDIF
		\ENDFOR
	\end{algorithmic}
\end{algorithm}

\section{Experiments}
\label{sec:exp}

\subsection{Experiment setup}

Our experiments are conducted on the large-scale ImageNet VID dataset ~\cite{russakovsky2015imagenetvid} (VID). The training, validation and testing sets of VID contain 3862, 555 and 937 video snippets, respectively. Training and evaluation are performed on the 3862 video snippets from the training set and the 555 snippets from the validation set, respectively. The frame rate is 25 or 30 fps for most snippets. The frame-level bounding box annotations for the training and validation sets are available for training. There are totally 30 categories of objects, which are a subset of the ImageNet DET dataset. Following the standard settings in ~\cite{zhu2017deep}, the training set includes both the VID training set and the ImageNet DET training set (only the same 30 categories are used), and the VID validation set is used for evaluating our method. All the frames are resized so that the shorter side has 600 pixels or the longer side has 1000 pixels, without changing aspect ratio for both training and testing. When reporting the results, we use the standard mean average precision (mAP) metric.

During training, each mini-batch consists of images from both ImageNet VID and ImageNet DET at the ratio of $2:1$. We use 4 Titan Xp GPUs for training. In ablation studies and comparisons with the state-of-the-art methods, we use a Titan Xp GPU and a Titan X GPU, respectively. SGD optimizer is used for training 120k iterations with a weight decay of 5e-4. The learning rates are $2.5\times10^{-4}$ and $2.5\times10^{-5}$ for the first 80K and the last 40K iterations, respectively. The ``simple version" of FlowNet ~\cite{dosovitskiy2015flownet} is pretrained on the Flying Chairs dataset. For clear comparison, no bells-and-whistles like multi-scale training/testing are used. For the rest experimental settings, we follow those of DFF ~\cite{zhu2017deep}.

\begin{table*}[htp]
		\caption{Ablation studies on ImageNet VID. The top part shows the selection of the studied methods. The bottom part shows the detection performance including both detection accuracy and speed. Note that the testing speed is obtained using a Titan Xp GPU, with a single frame as input. The entry marked with `*' means that Seq-NMS uses CPU for computation.}
	\vspace{1mm}
	\centering
	\begin{tabular}{l|cccccccc }
		\toprule
		Method           &  (a)    &  (b)    & (c)     & (d)     &     (e) &    (f)  & (g)  & (h) \\
		\midrule
		R-FCN  &\checkmark &\checkmark & \checkmark & \checkmark & \checkmark & \checkmark & \checkmark & \checkmark\\
		DFF          && \checkmark &&&&&         &       \\
		SFA - $\mathcal{W}\left( F'_t, M_{t+x} \right)$ in Eq.~\eqref{eq:sfa}&&& \checkmark & \checkmark&  \checkmark & \checkmark & \checkmark & \checkmark\\
		SFA - $\text{Conv}\left( R_{t+x} \right)$ in Eq.~\eqref{eq:sfa}&&& & \checkmark& \checkmark & \checkmark & \checkmark & \checkmark\\
		SFA - $\mathcal{N}_\text{tiny} D\left( I_{t+x} \right) $ in Eq.~\eqref{eq:sfa} &&&& & \checkmark& \checkmark & \checkmark & \checkmark \\
		DCN~ in $\mathcal{N}_\text{large}$   &&&&&  & \checkmark& \checkmark & \checkmark \\
		LFA - Eq.~\eqref{eq:lfa} &&&&&& & \checkmark& \checkmark  \\
		Seq-NMS   &         &&&&&         &         & \checkmark \\
		\midrule
		mAP (\%)         & 75.3    &  74.8  &  73.7  &  74.2    &  75.0   &  76.1   &  77.2   & 79.3  \\
		mAP (\%) (slow)  & 84.2    &  85.5  &  84.7  &  85.7    &  85.8   &  87.1   &  86.6   & 88.7\\
		mAP (\%) (medium)& 73.1    &  73.3  &  72.1  &  72.4    &  73.0   &  74.9   &  76.2   & 78.9  \\
		mAP (\%) (fast)  & 51.9    &  49.3  &  47.7  &  47.6    &  48.9   &  49.2   &  52.6   & 54.6   \\
		Speed (fps)       & 13      &  25   &  43  &  43      &  40     &    37   &  33     &  8*   \\
		\bottomrule
	\end{tabular}
	\vspace{3mm}
	\label{table:ablation}
\end{table*}

\subsection{Ablation studies}

We conduct ablation studies to validate the effectiveness of the studied components. All the ablation studies use the same hyper-parameters for both training and testing. We report the detection speed in term of frame per second (FPS) and the detection accuracy in mAP, mAP for slow moving objects, mAP for medium moving objects and mAP for fast moving objects, which are defined in ~\cite{zhu2017flow}. We first verify whether the motion vector from video compression matches the performance of FlowNet, and then check the performance improvement of the proposed tiny feature extraction network. Finally, we study the effectiveness of the long-term feature aggregation. Moreover, the results of adding convolution and SeqNMS are also provided. All results are listed in Table~\ref{table:ablation}. For these results, we provide the details and analyses for the methods (a)-(g) in the following.
\begin{itemize}
    \item [(a)] uses R-FCN with ResNet-101 to perform frame-by-frame object detection. The results are 75.3\% mAP at a speed of 13 FPS. It does not utilize any temporal information. This method is also called `frame baseline'.
    \item[(b)] is the DFF~\cite{zhu2017deep} method. DFF is also based on R-FCN with ResNet-101. It obtains 74.8\% mAP at speed of 25 FPS. Compared with the above frame baseline, the speed is significantly improved and the accuracy slightly drops.
    \item[(c)] means replacing the FlowNet in DFF with the motion vector extracted from compared video. It is defined in Eq.~\eqref{eq:sfa} as a part of the proposed short-term feature aggregation (SFA). The results show that the quality of motion vectors is lower than the flow computed using FlowNet. It has a 1.1\% mAP accuracy drop, but the speed is significantly improved, i.e., from 25 FPS to 41 FPS. In the following studies, the methods are based on the SFA framework.
    \item[(d)] adds the feature extracted form the residual error map using a convolution layer to remedy the accuracy drop of using motion vector. The results show that the residual error features give 0.5\% mAP improvement with negligible computation cost. Especially in the slow movement case, it can improve about 1\% mAP. 
    \item[(e)] adds the feature extracted from the downsampled original image using the tiny network $\mathcal{N}_\text{tiny}$. The tiny network gives an 0.8\% mAP improvement and its computation cost is only 3 FPS. During experiments, we find that the $3\times3$ convolution in $\mathcal{N}_\text{tiny}$ is important. If we do not add the $3\times3$ convolution, the result will drop to 74.3\% mAP, while adding the $3\times3$ convolution the result can be 75.0\% mAP. In the fast movement case, adding the tiny network can improve 1.3\% mAP. For the reason that when the object moves fast, the motion vectors and residual errors can not describe the movements accurately even the object is clear, then, the features extracted from the original can be helpful.
    \item[(f)] adds the deformable convolution ~\cite{dai2017deformable} in the backbone network for key frame feature extraction, i.e., $\mathcal{N}_\text{large}$. DCN improves 1.1\% mAP and costs the computation of 3 FPS. DCN is very helpful for slow and medium motion objects while improving slightly for fast moving objects. That means when the object moves fast, even a better feature extraction network can not get high-quality features. Thus adding temporal information is very important.
    \item[(g)] adds the long-term feature aggregation (LFA) defined in Eq~\eqref{eq:lfa}. LFA improves 1.1\% mAP and costs the computation of 4 FPS. Here, we reach the mAP of 77.2\% at the speed of 33 FPS on a Titan Xp GPU. It is the standard form of the proposed LSFA video object detector. We will use this detector to compare with the state-of-the-art methods. We can find that LFA can improve the accuracy of fast motion objects a lot. This long-term information alleviates the problems of motion blur, partial occlusion, viewpoint variation et al., yielding a more robust feature for the key frame. This shows that we can add more temporal information to improve the accuracy of fast motion. 
    \item[(h)] means using Seq-NMS as post-processing for the above detector in (f). The results show that detection accuracy of LSFA can be benefited from the Seq-NMS processing. The mAP boosts to 79.3\% mAP but the testing speed drops to 8 FPS with additional CPU computation. 
\end{itemize}

\begin{table*}[!htp]
  \caption{The detection accuracy and speed of the state-of-the-art video object detection methods and ours. For fair comparisons, all the following methods use ResNet-101 as backbone and all the values of testing speed are obtained using a Titan X GPU unless the RDN method ~\cite{deng2019relation} is tested on a TITAN V GPU. Results including FPS are borrowed from original papers. ``-" means the information is not available. }
  \vspace{1mm}
  \centering
  {\begin{tabular}{lcccc}
  \toprule
    Method &  Publication & Online & mAP (\%) & Speed (FPS) \\
    \midrule
     R-FCN ~\cite{dai2016rfcn} & NeurIPS 2016 & \cmark & 73.9 &  4.05 \\
     TPN ~\cite{hetang2017impression} & arXiv 2017 & \xmark & 68.4 & 2.1  \\
     DFF ~\cite{zhu2017deep} & CVPR 2017 & \cmark & 73.1 & 20.2   \\
     D\&T ~\cite{feichtenhofer2017detect} &  ICCV 2017 & \cmark    & 75.8 & 7.8  \\
     FGFA ~\cite{zhu2017flow} &  ICCV 2017 & \xmark  & 76.3 & 1.3\\
     SRFA ~\cite{zhu2018towards} & CVPR 2018 & \cmark  & 77.8 & 22.9\\
     ST-lattice ~\cite{chen2018optimizing} & CVPR 2018 & \xmark & 79.5 & 20\\
     MANet ~\cite{wang2018fully} & ECCV 2018 & \xmark & 77.6 & 7.8 \\
     PSLA ~\cite{guo2019pro} & ICCV 2019 & \cmark & 77.1 & 18.7 \\
     LRTR ~\cite{shvets2019leveraging} & ICCV 2019 & \xmark & 80.6  & 10 \\
     OGEMNet ~\cite{deng2019object} & ICCV 2019 & \cmark & 76.8 & 14.9 \\
     MMNet ~\cite{wang2018fodcv} & ICCV 2019 & \cmark & 73.0 & 41 \\
     SELSA ~\cite{wu2019sequence} & ICCV 2019 & \xmark & 80.3 & - \\
     RDN ~\cite{deng2019relation} & ICCV 2019 & \xmark & 81.8 & 10.6 (V) \\
     HQOL ~\cite{tang2020object} & TPAMI 2020 & \xmark & 80.6 & 2.8 \\
     MEGA ~\cite{chen20mega} & CVPR 2020 & \xmark & 82.9 & - \\
     LSTS ~\cite{jiang2020learning} & ECCV 2020 & \cmark &77.2 & 23.0\\
     LSFA (ours)    &  - &  \cmark  & 77.2 & 30  \\
     \bottomrule
  \end{tabular}}
  \label{table:compare}
\end{table*}

\subsection{Comparisons with state-of-the-art methods}

In Table~\ref{table:compare}, we compare state-of-the-art image- and video-based object detection methods to show a comprehensive landscape of object detection algorithms in the ImageNet VID task. The methods are listed in a chronological order. Their information of publication venues, supporting online inference or not, detection precision and speed are reported. All the methods use ResNet-101 as the backbone network. From the results, we can observe that the offline methods have better mAP but their inference speeds are low. The best mAP is obtain by MEGA ~\cite{chen20mega} published in CVPR 2020. The proposed LSFA method achieves 77.2\% mAP at 30 FPS using a Titan X GPU. There are two real-time methods, MMNet ~\cite{wang2018fodcv} and our LSFA; the mAP of LSFA is much higher than MMNet (77.2\% v.s. 73.0\%). Compared with the latest online video object detection method published in ECCV 2020, i.e. LSTS ~\cite{jiang2020learning}, our method has the same mAP but our inference speed is much faster.


\subsection{Detailed running time}

To have a deeper understanding on how does our method obtain real-time performance, we present the detailed running time of each individual component in Table~\ref{tab:runtime}. The results show that inference a key frame takes 151.28 ms while inference a non-key frame takes only 22.57 ms. The fast inference speed for non-key frame is achieved by the proposed short-term feature aggregation with the tiny feature extraction network. By setting 1 key frame with 11 non-key frames, our system obtains 33.3 per image in average (30 FPS). The upper speed bound of our system is 44.3 FPS (1000/22.57) and the lower speed bound is 6.6 FPS (1000/151.28). If the non-key frame interval becomes larger, the speed becomes higher and the mAP will become lower vice verse. Setting the interval to 11 can obtain a good trade-off. 

\begin{table}[!htp]
  \caption{Detailed running time for each individual component in our video object detection method tested on a Titan Xp GPU.}
  \label{tab:runtime}

\resizebox{\linewidth}{!}{
\begin{tabular}{cc|cc}
    \toprule
    \multicolumn{2}{c}{Key frame} & \multicolumn{2}{c}{Non-key frame} \\
    Component & Runtime (ms) & Component & Runtime (ms) \\ 
    \midrule
    ${N}_\text{large}$ & 103.55 & \multirow{2}{*}{${N}_\text{tiny}$} & \multirow{2}{*}{8.48}\\
    ${N}_\text{flow}$ & 29.89 & \\
    LFA & 9.57 & SFA & 5.82\\
    R-FCN head & 8.27 & R-FCN head & 8.27 \\
    Time/frame & 151.28 & Time/frame & 22.57\\
    \#Frames & 1 & \#Frames & 11 \\ 
    \midrule
    \multicolumn{2}{c}{Avg time for 12 frames} & \multicolumn{2}{c}{33.3 ms}\\
    \bottomrule
\end{tabular}
}
\end{table}

\subsection{Visualization}

\begin{figure*}[!htp]
    \centering
    \includegraphics[width=1.0\linewidth]{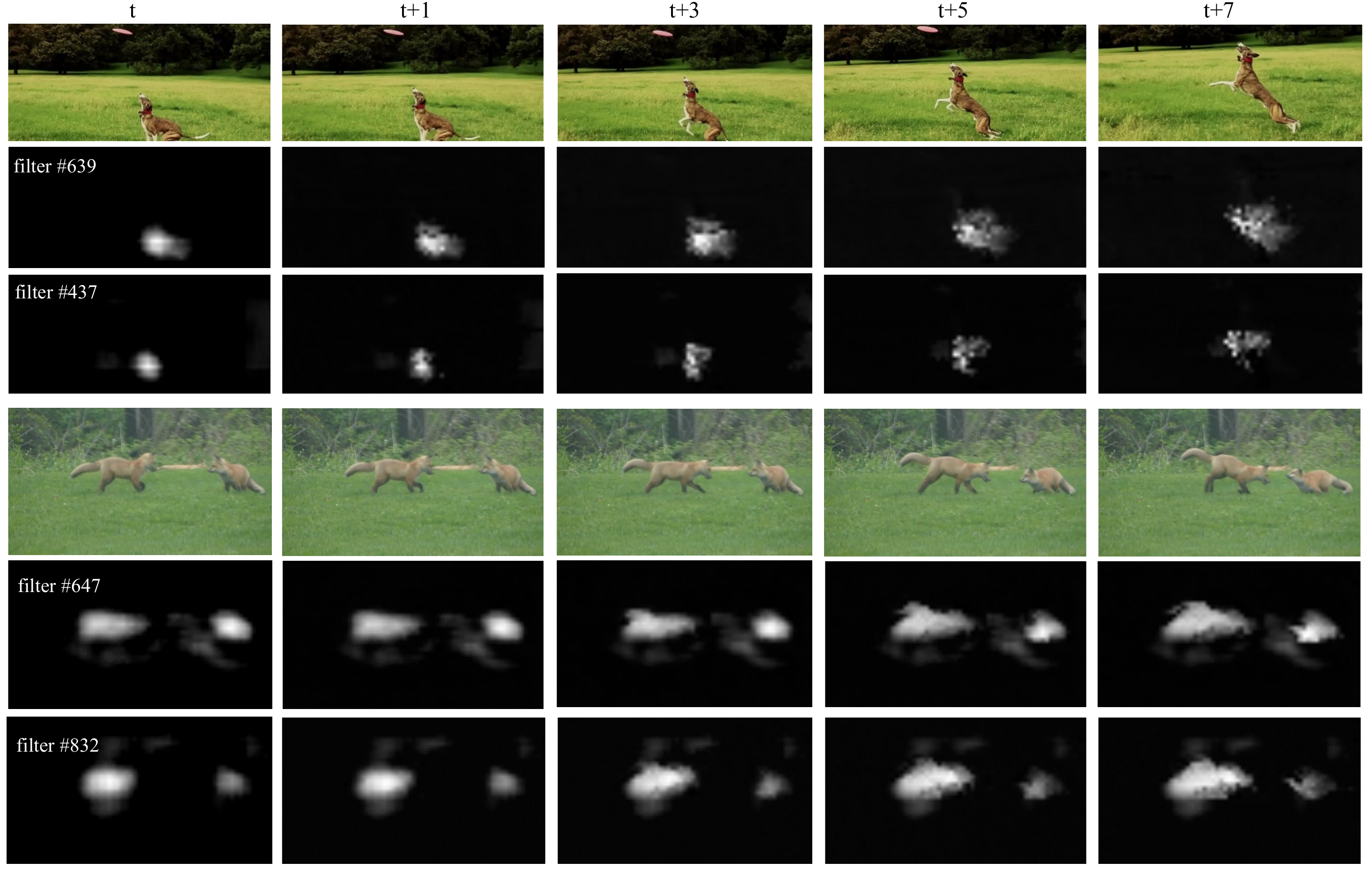}
    \caption{Feature visualization of the method (d) (i.e., $\mathcal{W}\left( F'_t, M_{t+x} \right) + \text{Conv}\left( R_{t+x} \right)$ in Eq.~\eqref{eq:sfa}) and method (e) (i.e., $\mathcal{N}_\text{tiny} D\left( I_{t+x} \right) $ in Eq.~\eqref{eq:sfa}). The first row and the fourth row are the original images. The second row and the fifth row visualize some features for method (e). The third row and the sixth row visualize some features for method (d). 
    $filters \#x$ means that the features from the $x$-th channel. Since the network of  method (d) and method (e) are independently trained, the same category object may stay in different channels. We can find that the visualization results from method (e) is better than method (d), the feature from method (e) are richer. }
    \label{fig:visualization}
\end{figure*}

We visualize some deep features that are fed into the R-FCN head network, i.e., the feature represented by the orange blocks in Figure~\ref{fig:network}. The purpose of this visualization is to study the effectiveness of the tiny feature extraction network in short-term feature aggregation, as Table~\ref{table:ablation} shows that this network yields higher accuracy in fast motion ($47.6\rightarrow48.9$). As shown in Figure~\ref{fig:visualization}, when the motion of object is fast, the propagated feature guided by the motion vectors plus the extracted feature from the residual errors fail to describe the movements of the object accurately, even if the object is clear. In such case, the features extracted from the original image tends to be helpful. Therefore, the visualization results provide a reasonable explanation for the effectiveness of the tiny network.

\section{Conclusion}
\label{sec:con}

In this paper, we proposed a high performance method, named Long Short-Term Feature Aggregation (LSFA), for video object detection. Based on the deep feature flow method, we use motion vectors and residual errors to replace the FlowNet in a simple way, which can achieve similar accuracy with FlowNet but faster speed. Then we add a tiny network to obtain overall information of the non-key frame to improve the quality of features for recognition. Experiment results on the large-scale ImageNet show that our method achieves both high accuracy and real-time speed, which are better or on par with the state-of-the-art methods. In the future, we would like to explorer effective attention ~\cite{huang2020ccnet} and weakly-supervised learning ~\cite{tang2020pcl} for video object detection.

{\small
\bibliographystyle{ieee_fullname}
\bibliography{cvpr}
}

\end{document}